\documentclass[10pt,twocolumn,letterpaper]{article}

\usepackage{iccv}
\usepackage{times}
\usepackage{epsfig}
\usepackage{graphicx}
\usepackage{amsmath}
\usepackage{amssymb}
\usepackage{multirow}
\usepackage{subfig}

\usepackage[pagebackref=true,breaklinks=true,letterpaper=true,colorlinks,bookmarks=false]{hyperref}

\iccvfinalcopy 

\def\iccvPaperID{1691} 

\ificcvfinal\fi
\begin{document}

\title{Large Scale Business Store Front Detection from Street Level Imagery}

\author{Qian Yu, Christian Szegedy, Martin C. Stumpe,\\ Liron Yatziv, Vinay
Shet, Julian Ibarz, Sacha Arnoud\\
Google StreetView
\\
{\tt\small  qyu, szegedy, mstumpe, lirony, vinayshet, julianibarz, sacha@google.com}
}

\maketitle

\begin{abstract}   
We address the challenging problem of detecting business store
fronts in street level imagery. Business store fronts are a challenging class
of objects to detect due to high variability in visual appearance. Inherent
ambiguities in visually delineating their physical extents, especially in
urban areas, where multiple store fronts often abut each other, further increases
complexity. We posit that traditional object detection approaches such as
those based on exhaustive search or those based on selective search followed
by post-classification are ill suited to address this problem due to these
complexities. We propose  the use of a Multibox~\cite{mbox_cvpr_2014}
based approach that takes as input image pixels and directly outputs store
front bounding boxes. This end-to-end learnt approach instead preempts the need for
hand modelling either the proposal generation phase or the post-processing
phase,  leveraging large labelled training datasets. We demonstrate
our approach outperforms the state of the art detection techniques with a
large margin in terms of performance and run-time efficiency. In the evaluation, we show this approach achieves human accuracy in the low-recall settings. We also provide
an end-to-end evaluation of business discovery in the real world.
\end{abstract}
\section{Introduction}
\begin{figure}[t]
\begin{center}
   \includegraphics[width=1\linewidth]{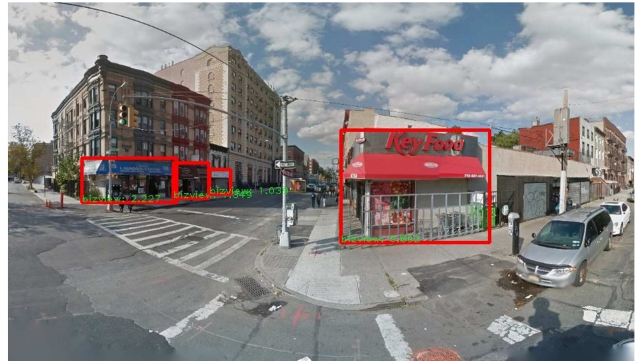}
\end{center}
   \caption{Typical Street View image showing multiple store fronts. The
red boxes show successfully detected store fronts using the 
   approach presented in this paper}
\label{fig:intro}
\end{figure}

The abundance of geo-located street level photographs available on the internet
today provides a unique 
opportunity to detect and monitor man-made structures to help build precise
maps. One example of such 
man-made structures is local businesses such as restaurants, clothing stores,
gas stations, pharmacies, 
laundromats, etc. There is a high degree of consumer interest in searching
for such  businesses 
through local queries on popular search engines.
Accurately identifying the existence of such 
local businesses worldwide is a non-trivial task. We attempt to automatically
identify a 
business by detecting its presence on geo-located street level photographs.
Specifically, we explore the 
world's largest archive of geo-located street level photos, Google Street
View~\cite{vincent_taking_2007,anguelov_google_2010}, to extract business
store fronts. Figure~\ref{fig:intro} illustrates 
our use case and shows sample detections using the approach presented in
this paper. 

\begin{figure}[tb]
\begin{center}
\subfloat[gas station]{
   \includegraphics[height=1in]{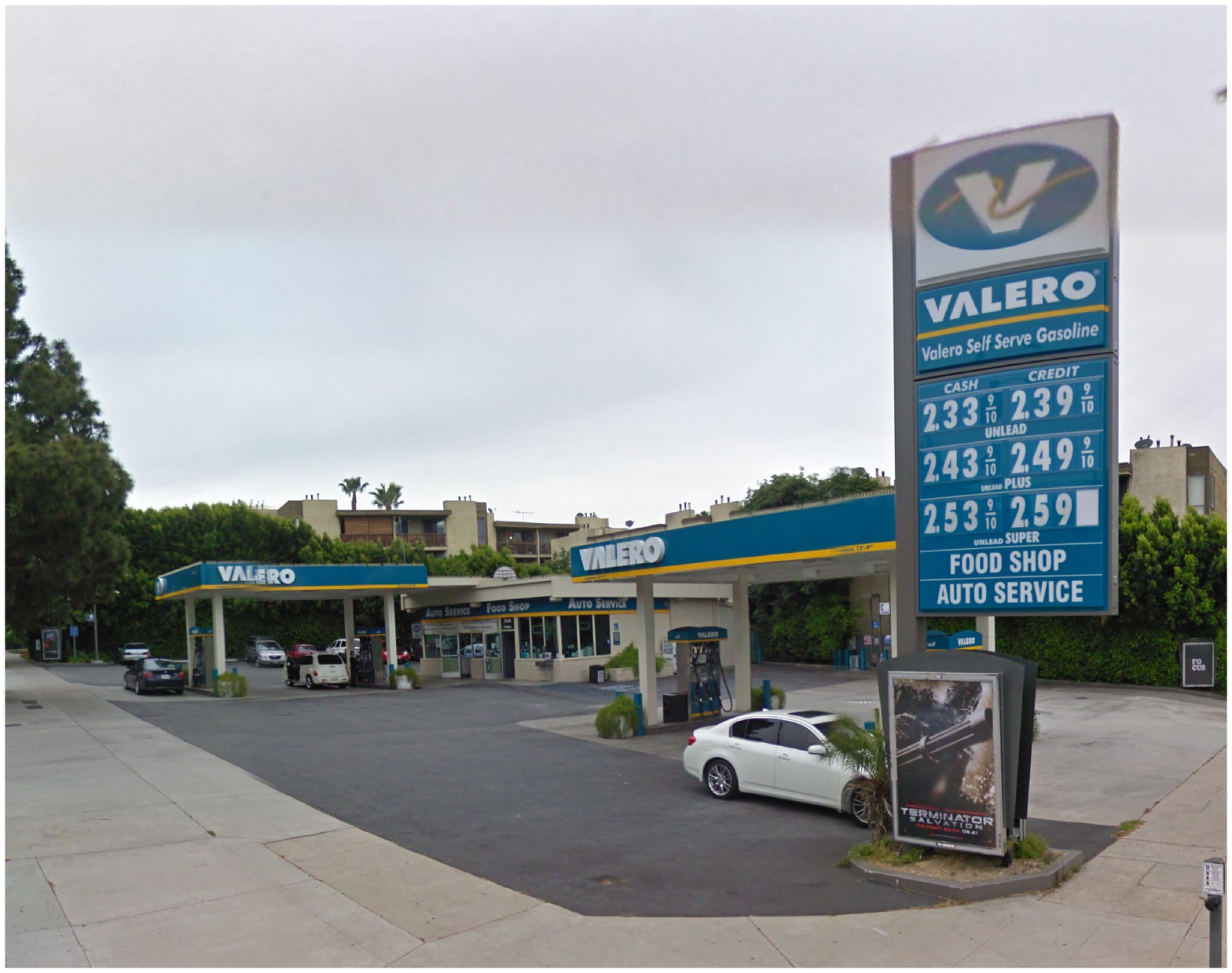}
   }
\subfloat[hotel]{
    \includegraphics[height=1in]{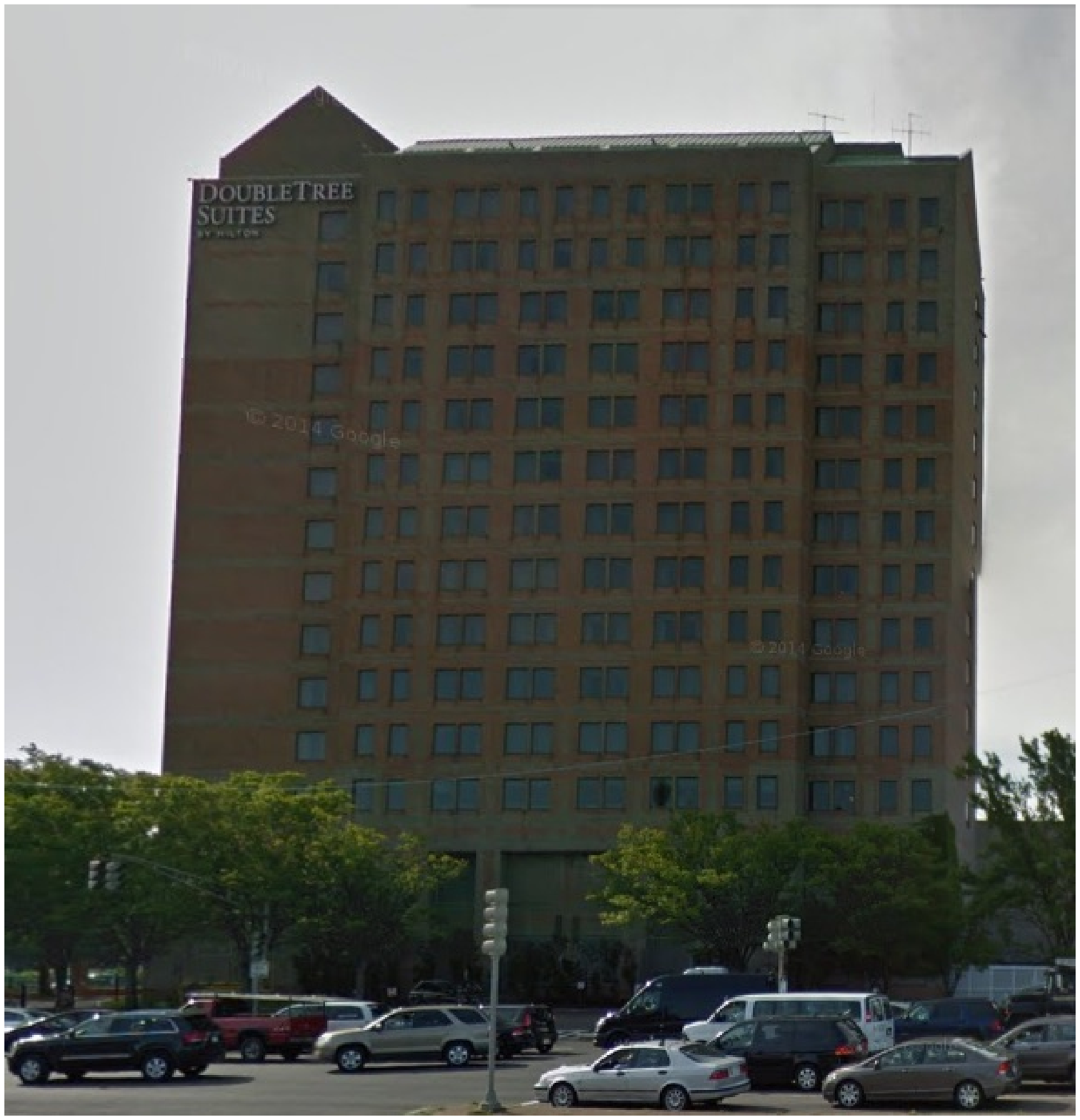}
   }
\subfloat[dry cleaner store]{
    \includegraphics[height=1in]{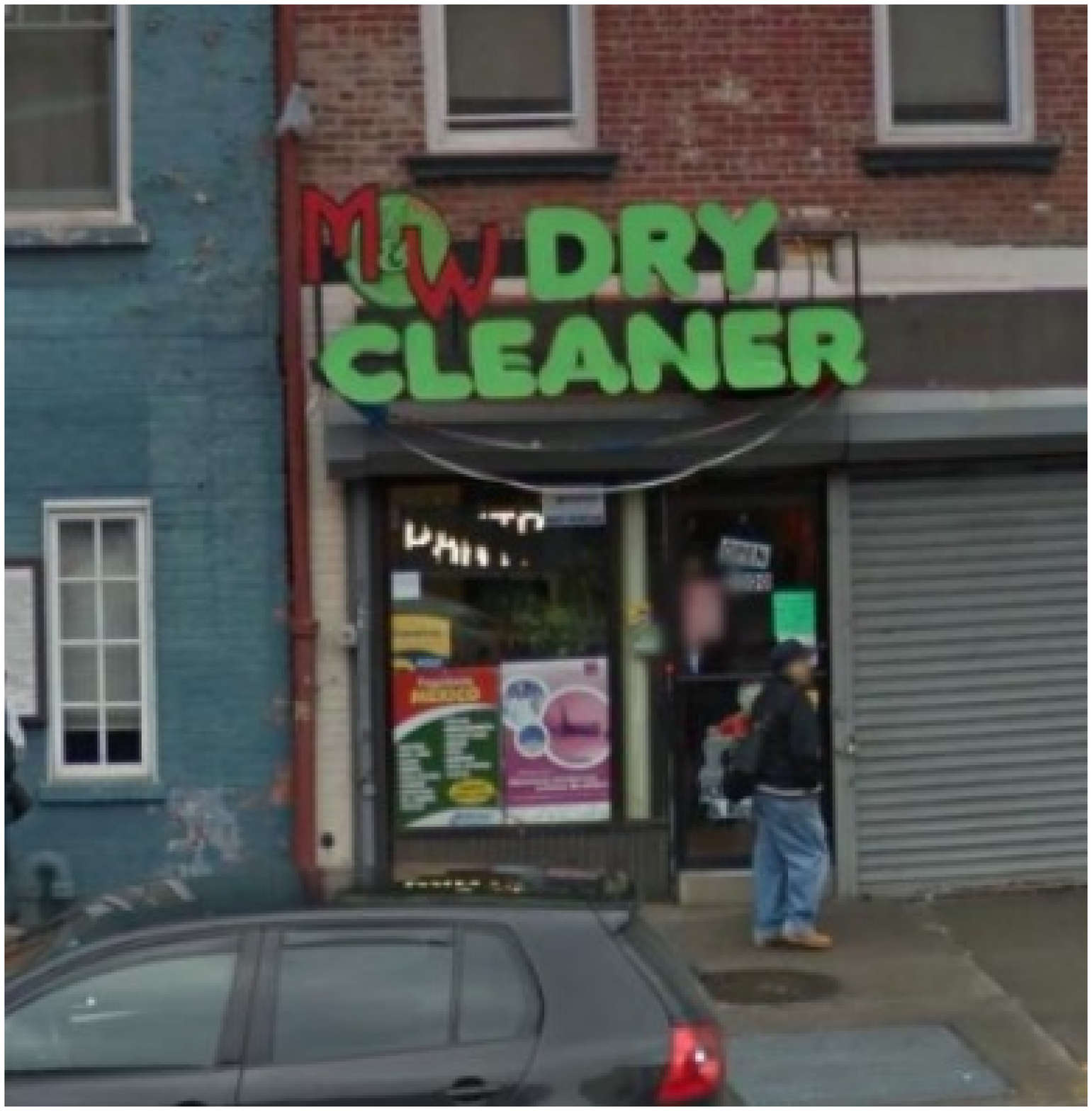}
   }\\
\subfloat[local store in Japan]{
   \includegraphics[height=1.02in]{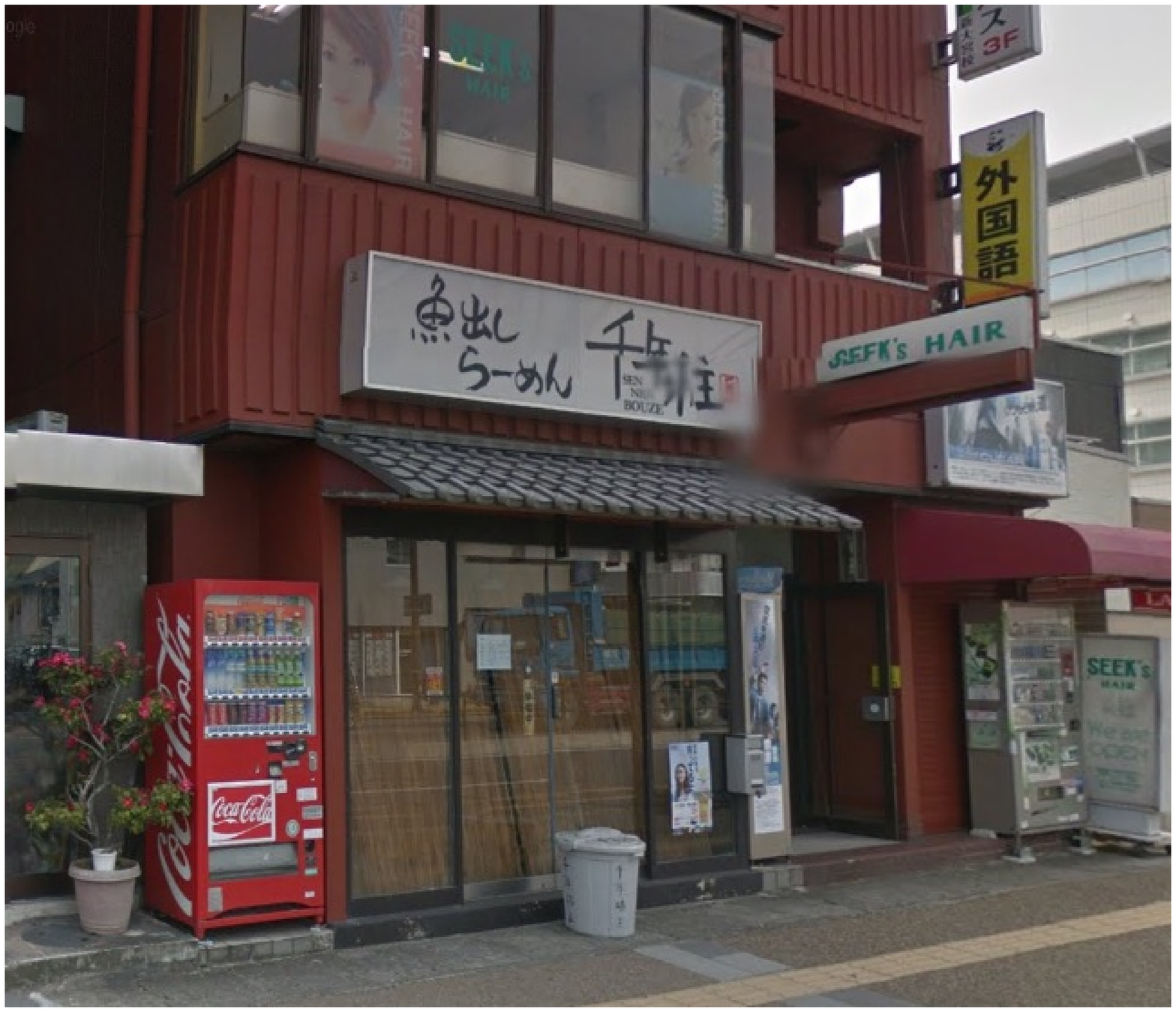}
   }
\subfloat[Several businesses together]{
   \includegraphics[height=1.02in]{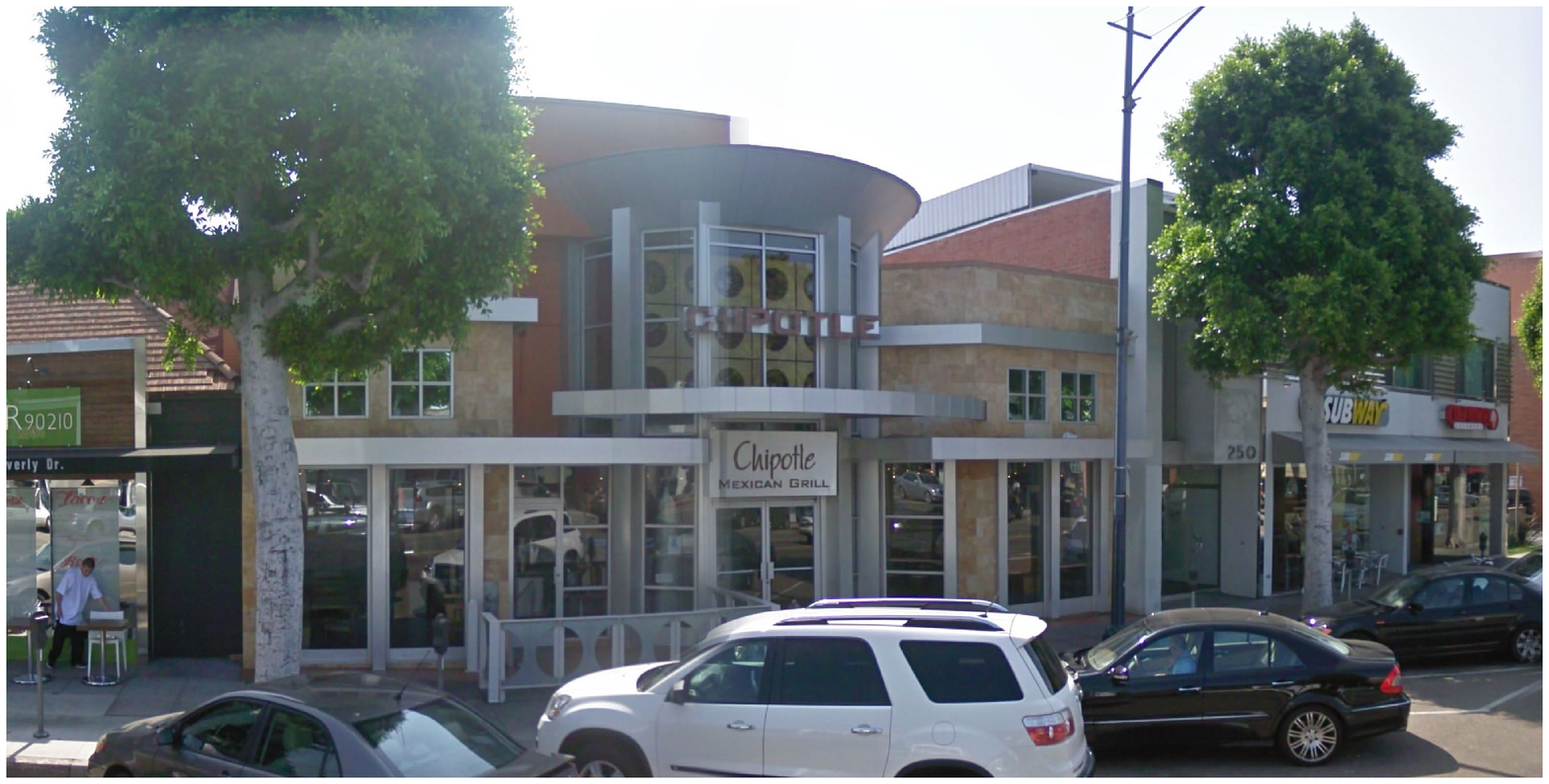}
}
\end{center}
   \caption{Precisely detecting business is a challenging task. The top row
illustrates the large variance between different categories. The bottom row
shows business boundary is difficult to precisely define.  }
\label{fig:challenges}
\end{figure}

Extracting arbitrary business store fronts from Street View imagery is a
hard problem. Figure~\ref{fig:challenges} illustrates some of the challenges.
The complexity comes from the high degree of intra-class variability in the
appearance of store fronts across business categories and geographies (Figure~\ref{fig:challenges}
a-d), 
inherent ambiguity in the physical extent of the store front (Figure~\ref{fig:challenges}
d-e), businesses abutting each other in urban areas, 
and the sheer scale of the occurrence of store fronts worldwide (likely in
the hundreds of millions). These factors make this an ambiguous task even
for 
human annotators. Image acquisition factors such as noise, motion blur, occlusions,
lighting variations, specular reflections, perspective, 
geo-location errors, \etc. further contribute to the complexity of this problem.
Given the scale of this problem and 
the turn over rate of businesses, manual annotation is prohibitive and unsustainable.
For automated approaches, runtime efficiency is 
highly desirable for detecting businesses worldwide in a reasonable time-frame.

Detecting business store fronts is the first and most critical step in a
multi-step process to extract usable 
business listings from imagery. Precise detection of store fronts enables
further downstream processing 
such as geo-location of the store front, extraction of business names and
other attributes, e.g. category classification. 
In this paper, we focus on this critical first step, namely, precisely detecting
business store fronts at a large 
scale from StreetView imagery.

In this paper, we propose a Convolutional Neural Network (CNN) based approach
to large scale business store front detection. 
Specifically, we propose the use of the MultiBox~\cite{mbox_cvpr_2014} approach
to achieve this goal. The Multibox approach 
uses a single CNN to take  image pixels as input and directly predict bounding
boxes, corresponding
to the object of interest, together with their confidences. The inherent
ambiguities in delineating store fronts and their tendency 
to abut each other in urban areas provides a challenge to traditional object
detection approaches such as those based on exhaustive 
search or those based on selective search followed by a post-classification
stage. In this paper, we have a comparative study to show that the end-to-end
fully learned Multibox approach outperforms traditional object detection
approaches both in accuracy and in run-time efficiency,
enabling automatic business store front detection at scale.

In our comparison with other two approaches, Selective Search \cite{Uijlings13}
and Multi-Context Heatmap \cite{hydra}, we found that the head on approach
of Multibox attacking the detection problem directly improves the quality
of results while reducing the engineering effort. Selective search is designed
specifically for natural objects, so its coverage is inferior for store fronts
which require very subtle visuals cues to be separated. On the other hand,
heatmap based approaches like \cite{deep_object} and \cite{hydra} do not
produce bounding boxes directly, but need to post-process some intermediate
representation: heatmaps produced by convolutional networks. This incurs
extra engineering effort in the form of an additional step that needs to
be optimized for new use cases. In contrast, MultiBox learns to produce the
end result directly by the virtue of a single objective function that can
be optimized jointly with the convolutional features. This eases adaptation
to a specific domain. Also, the superior quality of the solution comes with
a significant reduction of computational cost due to extensive feature sharing
between the confidence and regression computations for a large number of
object proposals simultaneously.

\section{Related Work} 

The general literature on image understanding is vast. Object classification
and detection \cite{lsvm-pami} has been driven by the Pascal VOC object detection
benchmark [8] and more recently the ImageNet Large Scale Visual Recognition
Challenge (ILSVRC) \cite{pascal-voc-2011}. Here, we focus on reviewing related
work on analyzing Street View data, object detection and the use of Deep
Convolutional Networks.

\noindent\textbf{Analyzing Street View Data}.
Since its launch in 2007, Google Street View \cite{vincent_taking_2007,anguelov_google_2010}
has been used by the computer vision community as both a test bed for algorithms~\cite{lee_style-aware_2013,xiao_multiple_2009}
and a source from which data is extracted and analyzed~\cite{goodfellow_multi-digit_2013,zamir_accurate_2010,micusik_piecewise_2009,cornelis_3d_2008}.

Early work on leveraging street level imagery focused on 3D reconstruction
and city modeling, such as in \cite{cornelis_3d_2008,micusik_piecewise_2009}.
Later works have focused on extracting knowledge from Street View and leveraging
it for particular tasks. In~\cite{zamir_accurate_2010} the authors presented
a system in which SIFT descriptors from $100,000$ Street View images were
used as reference data to be queried upon for image localization. Xiao \etal~\cite{xiao_multiple_2009}
proposed a multi view semantic segmentation algorithm that classified image
pixels into high level categories such as ground, building, person, etc.
Lee et al.\ \cite{lee_style-aware_2013} described a weakly supervised approach
that mined mid-level visual elements, and their connections in geographic
datasets. Most similar to our work, is that of Goodfellow et al.\ \cite{goodfellow_multi-digit_2013}.
Both work utilizes Street View as a map making source, and data mine information
about real world objects. They focus on understanding street numbers, while
we are concerned with local businesses. They specifically describe a method
for street number transcription in Street View data.
Their approach unified the localization, segmentation, and recognition steps
by using a Deep Convolutional Network that operates directly on image pixels.
Their method, which was evaluated on tens of millions of annotated street
number images from Street View, achieved above 90\% accuracy and was comparable
to human operator precisions at a coverage above 89\%.

\noindent\textbf{Convolutional Networks}.
Convolutional Networks~\cite{fukushima_neocognitron:_1980, lecun_gradient-based_1998}
are neural networks that contain sets of nodes with tied parameters. Increases
in size of available training data and availability of computational power,
combined with algorithmic advances such as piecewise linear units~\cite{jarrett_what_2009,goodfellow_maxout_2013}
and dropout training~\cite{hinton_improving_2012} have resulted in major
improvements in many computer vision tasks. Krizhevsky et al.\ \cite{krizhevsky_imagenet_2012}
showed a large improvement over the state of the art in object recognition.
This was later improved upon by Zeiler and Fergus~\cite{zeiler_visualizing_2013},
and Szegedy et al.\ \cite{szegedy_going_2014}.

On immense datasets, such as those available today for many tasks, overfitting
is not a concern; increasing the size of the network provides gains in testing
accuracy. Optimal use of computing resources becomes a limiting factor. To
this end Dean et al.\ developed DistBeleif~\cite{dean_2012}, a distributed,
scalable implementation of Deep Neural Networks. We base our system on this
infrastructure.

\noindent\textbf{Object Detection}.
Traditionally, object detection is performed by exhaustively searching for
the object of interest in the image. Such approaches produce a probability
map corresponding to the existence of the object at a location. Post-processing
of this probability map, either through non-maxima suppression or mean-shift
based approaches, then generates discrete detection results. To counter the
computational complexity of exhaustive search, selective search~\cite{Uijlings13}
by Uijlings et al. uses image segmentation techniques to generate several
proposals drastically cutting down the number of parameters to search over.
Girshick et al proposed R-CNN~\cite{girshick2014rcnn} which uses a convolutional
post-classifier network to assign the final detection scores. The MultiBox
by Erhan at al.~\cite{mbox_cvpr_2014} takes this approach even further by
adopting a fully learnt approach from pixels to discrete bounding boxes.
The end-to-end learnt approach has the advantage that it integrates the proposal
generation and post-processing using a single network to predict a large
number of proposals and confidences at the same time. Although MultiBox can
produce high quality results by relying on the confidence output of the MultiBox
network alone, the precision can be pushed further by running extra dedicated
post-classifier networks for the highest confidence proposals. Even with
the extra post-classification stage, the MultiBox approach can be orders
of magnitude faster than R-CNN depending on the desired recall.

\section{Proposed Approach}

Most state-of-the-art object detection approaches utilize a proposal generation
phase followed by a postclassification pass. In our case,
traditional hand-crafted saliency based proposal generation methods have two
fundamental issues. First of all, our images are very large and detailed, so
we end up with a very large number (on average 4666)
of proposals per panorama. This makes the postclassification pass very expensive
computationally at the required scale. Secondly, the
coverage of the selective search \cite{Uijlings13} based proposals at $0.5$ overlap
is only 62\%. We hypothesize that this is due to the fact that the boundaries
between businesses require the utilization of much more subtle clues to be
separated from each other than large, clearly disjoint natural objects.

The large amount of training data makes this task a prime candidate
for the application of some learned proposal generation approach.
MultiBox was introduced in \cite{mbox_cvpr_2014} and stands out for
this task given its relative modest computational cost and its high detection
quality on natural images \cite{mbox_2015}.

\subsection{MultiBox}
\label{MultiBox}
The general idea of MultiBox \cite{mbox_cvpr_2014} is to use a single
convolutional network evaluation to directly predict multiple bounding box
candidates together with their confidence scores. MultiBox achieves this
goal by using a precomputed clustering of the space of possible object
locations into a set of $n$ ``priors''. The output of the convolutional network
is $5n$ numbers: the $4n$-dimensional ``location output'' (four values
for each prior) and the $n$-dimensional ``confidence output'' (one value for
each prior).
These $5n$ numbers are predicted by a linear layer fully connected to the
$7\times 7$ grid of the Inception module
described in \cite{mbox_2015} for which the filter sizes of that module are reduced to 64 (32 in the 1x1 and 3x3 convolutional layers each). 
This is necessary to constrain the number of parameters; for example, 12.5 million for $n=800$ priors. 

The priors have double purpose. During training time, the priors are matched
with the ground-truth boxes $g_j$ via maximum weight matching, where the
edge weight between box $g_j$ and prior $p_i$ is their Jaccard overlap.
Let $(x_{ij})$ denote the adjacency matrix of that matching:
that is $x_{ij}=1$ if ground-truth box $j$ was matched with prior $i$.
$x_{ij}=0$ for all other pairs of $(i, j)$. Note that $x_{ij}$ is independent
of the network prediction, it is computed from the ground-truth locations
and priors alone.
During training, the location output $l_i'$ of the network for
slot $i$ (relative to the prior) should match the ground-truth box $g_j$
if the ground-truth box $g_j$ was matched with prior $i$ (that is if $x_{ij}=1$).
Since the network predicts the location $l_i'$ relative to $i$-th prior,
we set $l_i = l_i' + p_i$ which is the prediction of ground-truth
location $g_j$, when $i$ is matched with $j$.
The target for the logistic confidence output $i$ is $1$ in this case and
is set to $0$ for all priors that are not matched with any ground-truth box.
The overall MultiBox loss is then given by:
$$ \sum_{i,j}x_{ij}\left(\frac{\alpha}{2}\|l_i-g_j\| - log(c_i)\right) - \sum_i(1-\sum_jx_{ij})log(1-c_i), $$
which is the weighted sum of the $L_2$ localization loss and the logistic
loss for the confidences. We have tested various values of $\alpha$ on
different data-sets and based on those experiments, we have used
$\alpha=0.3$ in this setup as well.
%

This scheme raises the question of whether we should pick specialized
prior for this task. However, we found that any set of prior that
covers the space of all rectangles (within a reasonable range of size and
aspect ratios) results in good models when used for training. Therefore, in our
setting, we reused the same set of $800$ priors that were
derived from clustering all the objects in the ILSVRC 2014 dataset.
Given the qualitatively tight inferred bounding boxes from our results,
we do not expect significant gains from using a different set of priors
specifically engineered for this task.

Furthermore, the fact that bounding boxes of businesses do not tend
to overlap means that the danger of two store fronts matching the exact
same prior is much lower than for natural objects that can occur in
cluttered scenes in highly overlapping positions. Also the low probability of
overlap allows for applying non-maximum suppression with a relatively low
overlap threshold of 0.2. This cuts the number of boxes that need to
be inspected in the post-classification pass significantly.

The quality of MultiBox can be significantly enhanced by applying it in
a very coarse sliding window fashion: we have used three scales
with a minimum overlap of 0.2 between adjacent tiles, which ends up with
only 87 crops in total for an entire panorama.
In the following, this approach will be referred to as
\emph{multi-crop evaluation}.
For detecting objects in natural web images, single crop
evaluation works well with MultiBox,
but since our panoramas are high resolution, smaller businesses
cannot be reliably detected from a low resolution version of a single panorama.
However, if the proposals coming from the various crops are merged without
postprocessing, businesses not fully contained in one crop tend
to get detected with high confidence and suppress the more complete views
of the same detection.
To combat this failure mode, we need to drop every detection that abuts one of the
edges of the tile,
unless that side also happens to be a boundary of the whole panorama. After multi-crop evaluation, we first drop the proposals
that are below a certain threshold and then drop the ones
that are not completely contained in the (0.1, 0.1) - (0.9,
0.9) subwindow of the crop. A non-maximal suppression
is applied to combine all of the generated proposals. There
is not any preprocessing in terms of geometry rectification
or masking out sky or ground regions. 

\subsection{Postclassification}
We found that postclassification could increase the average precision of
the detections by 6.9\%. For this reason, we trained a
GoogLeNet~\cite{szegedy_going_2014} model and applied it in
the R-CNN manner described in \cite{girshick2014rcnn}: i.e.
extending the proposal by $16.6$\% and
applying an affine transformation it to the $224\times 224$ receptive
field of the network.
For any given box in an image $I$, let $B(b)$ denote the event that box
$b$ overlaps the bounding box of a business with at least 0.5 Jaccard overlap.
Our task is to estimate $P(B(b)|I)$ for all proposals produced by the MultiBox.
This probability can be computed by marginalizing over each detection 
$b_i$ that has at least $0.5$ overlap with $b$:
$$P(B(b)|I) = \sum_{b_i} P(B(b)|D(b_i))P(D(b_i)|I),$$ where $D(b_i)$ is the event
that multibox detects $b_i$ in the image. This suggests that the probability
that box $b$ corresponds to a business can be estimated by a sum
of products of the confidence scores $\sum S_p(b)S_d(b_i)$
with all boxes $b_i$ overlapping $b$ with Jaccard similarity greater than
$0.5$, where $S_p$ and $S_d$ denotes the score of postclassifer and that of the
MultiBox detector network, respectively. In practice however,
we used non-maximum suppression with the very low threshold of 0.2 on
top of the detected boxes leaving us with a single term in the above sum:
$S_p(b)S_d(b)$, which is simply the product of the scores of the
MultiBox network and the postclassifier network for box $b$.
This is in fact the final score used for ranking our
detections at evaluation time.

\subsection{Training Methodology}
Both the MultiBox and postclassifier networks were trained with the
DistBelief \cite{dean_2012} machine learning system using stochastic gradient
descent. For Panorama images, we downsized the original image
by a factor of 8 for training both the MultiBox and the post-classifier
networks. The postclassifier network was
trained with random mixture of positive and negative crops in 1:7 ratio.
The negative crops were generated with MultiBox output with a low
confidence threshold.
\subsection{Detection in Panorama Space}
\begin{figure}[htb]
\begin{center}
   \includegraphics[width=3.3in]{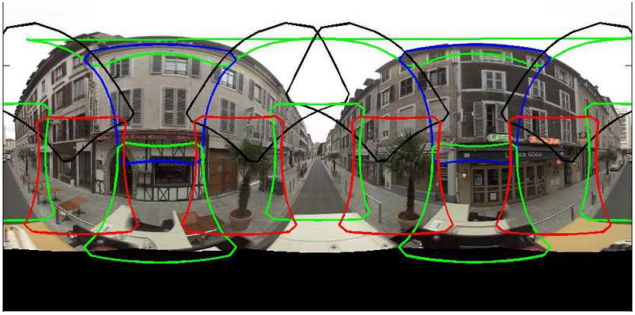}
\end{center}
   \caption{StreetView panoramas are composed of multiple individual images
(outlined in colors) which are projected to a sphere and blended together represented as a 2D equirectangular projection.}
\label{fig:pano-space}
\end{figure}

To avoid the loss of recall due to restriction of field of view, we
detect business store fronts in panorama space. Street View panoramas are created
by projecting individual camera images onto a sphere and stitching them together.
The resulting panorama image is represented in equirectangular projection,
i.e. the spherical image is projected onto a plane with 360$^{o}$ along the
horizon and 9$0^{o}$ up and 90$^{o}$ down as shown in Figure \ref{fig:pano-space}.
Each camera image only has a relatively small field-of-view, which makes business
detection
in the individual camera images not feasible since camera images often cut through
store fronts
as shown in Figure \ref{fig:pano-space}. Hence, our approach is trained and tested on
the equirectangular panoramas. As compared to single camera images, this
image representation has the disadvantage of having the equirectangular distortion
pattern.  Experiments show that the Deep Convolutional Network is
 able to learn the store front detection even under this distortion.  

\section{Results}
In this section we present our empirical results. First, we describe how
training and testing datasets are prepared. Next, we present the evaluation
procedure and compare our approach with other state of the art object detection
approaches. Finally, we have an end-to-end evaluation of the overall business
discovery results in the real world. Some qualitative detection results in
panorama space are shown at the end of paper.

\subsection{Business Store Front Dataset}
There is no large scale business store front dataset available publicly.
We have labeled about 2 million panorama images in more than 12 countries.
Annotations for this dataset were done through a crowd-sourcing system. The
original resolution of panorama image is 13312x6656. For most of the business
store fronts, the width varies from 200 to 2000 and the aspect ratio varies
from 1/5 to 5/1. 

Since businesses can be imaged multiple times from different angles, the
splitting between training and testing  is location aware, similar to the
one used in \cite{bizclassify}.  This ensures that businesses in the test
set were never observed in the training set. 

Similar to most object annotation tasks, it is hard to enforce the completeness
of the annotation, especially at this scale. In order to have a proper evaluation
on this problem, we sub-sampled a smaller test dataset with 2,000 panorama
images, where we enforce the completeness of annotations by increasing
operator replication and adding a quality control stage in the crowd-sourcing
system to ensure  all
visible businesses  from the panoramas are labeled with
the best effort. Compared to the original 2934 store front annotations in
the original test set, about 11,000 annotations were created. This indicates
how incomplete the training dataset is. We use this smaller but more complete
dataset as the test set for comparison.

\subsection{Runtime Quality Tradeoff }
\begin{figure*}[htb]
\begin{center}
\subfloat[AP increases as \# of crops increases.]{
   \includegraphics[height=1.6in]{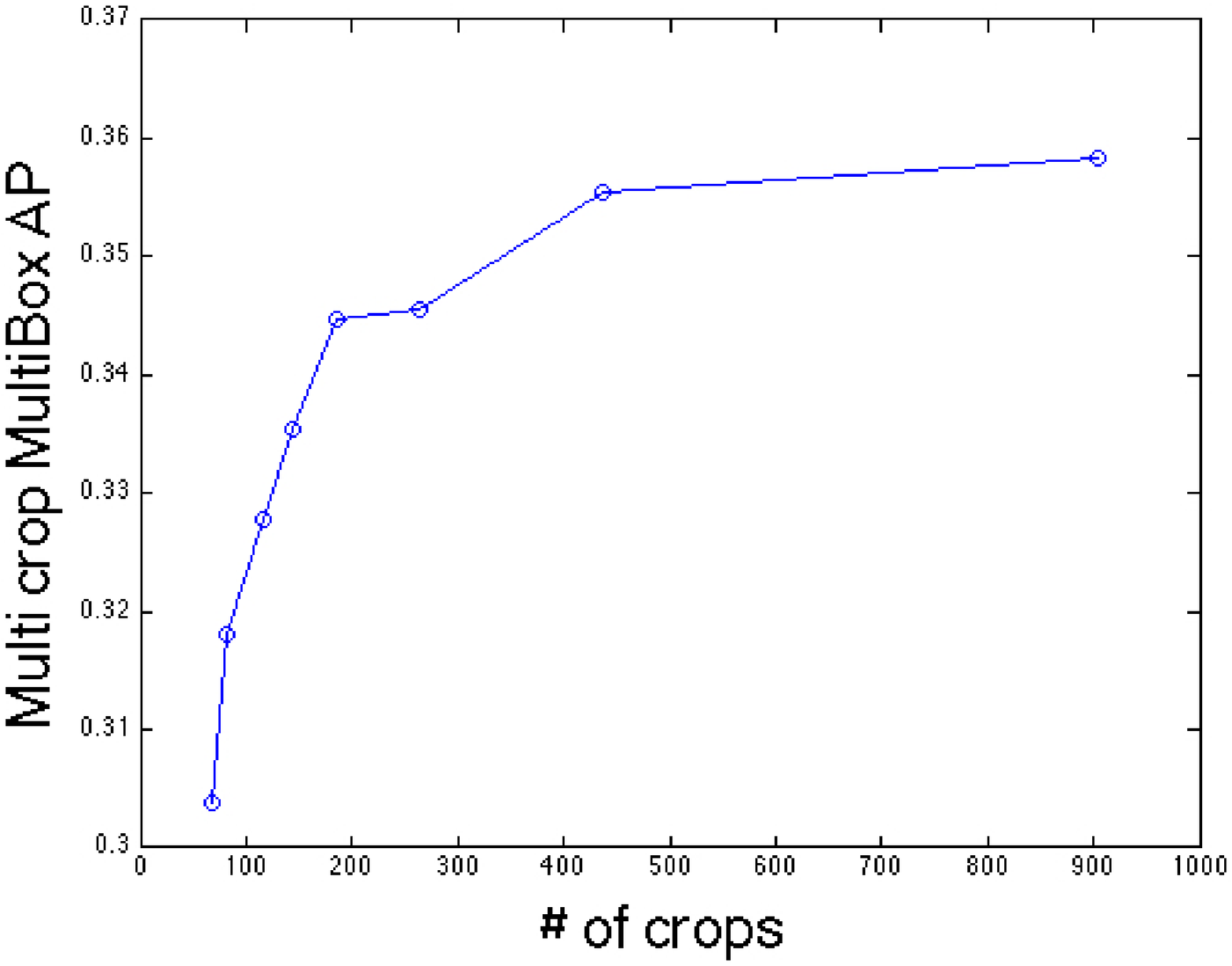}
}
\subfloat[Improvement is mostly due to better
recall.]{
   \includegraphics[height=1.7in]{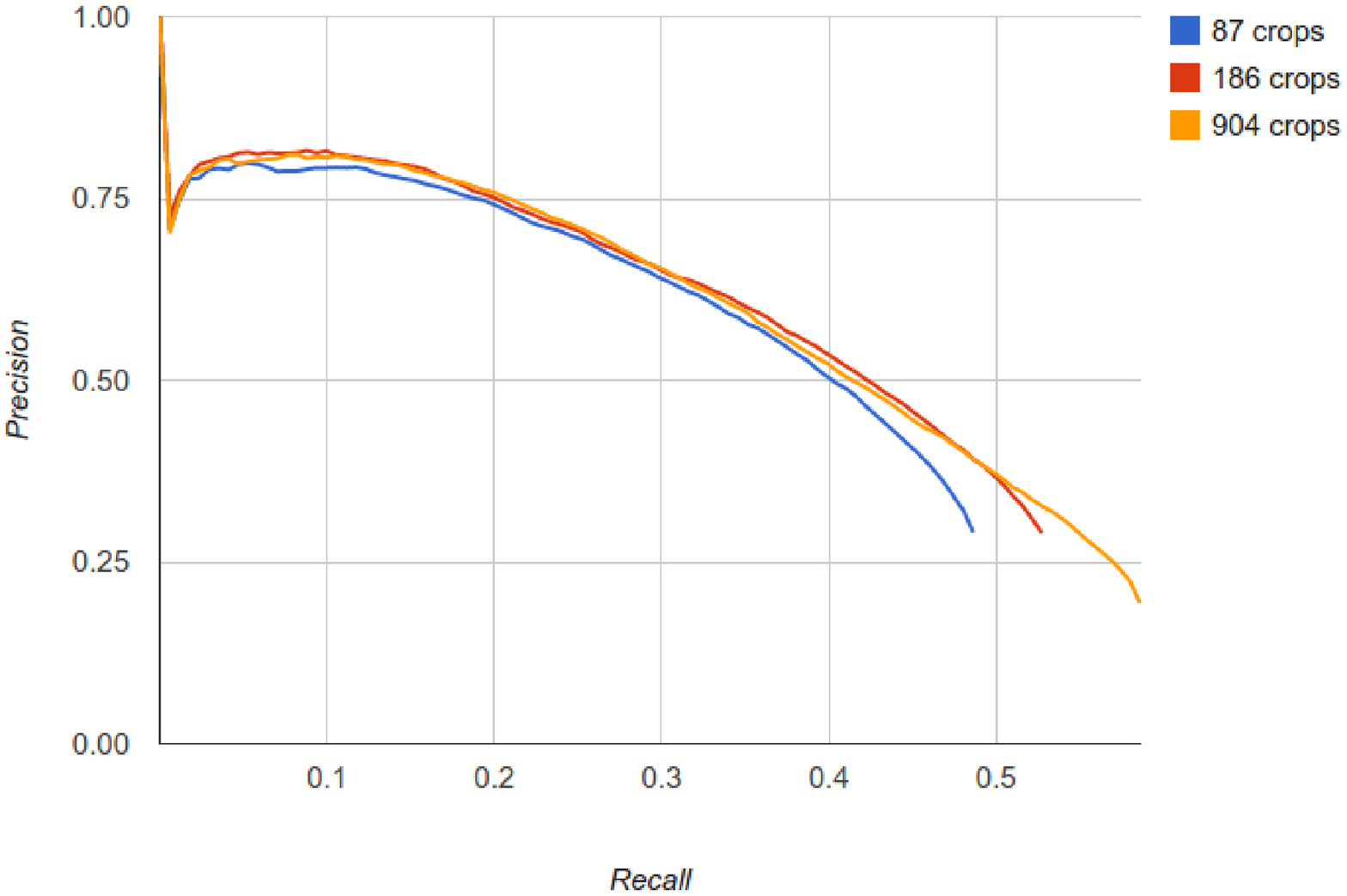}}
\subfloat[AP varies according to \# of proposals]{   
   \includegraphics[height=1.6in]{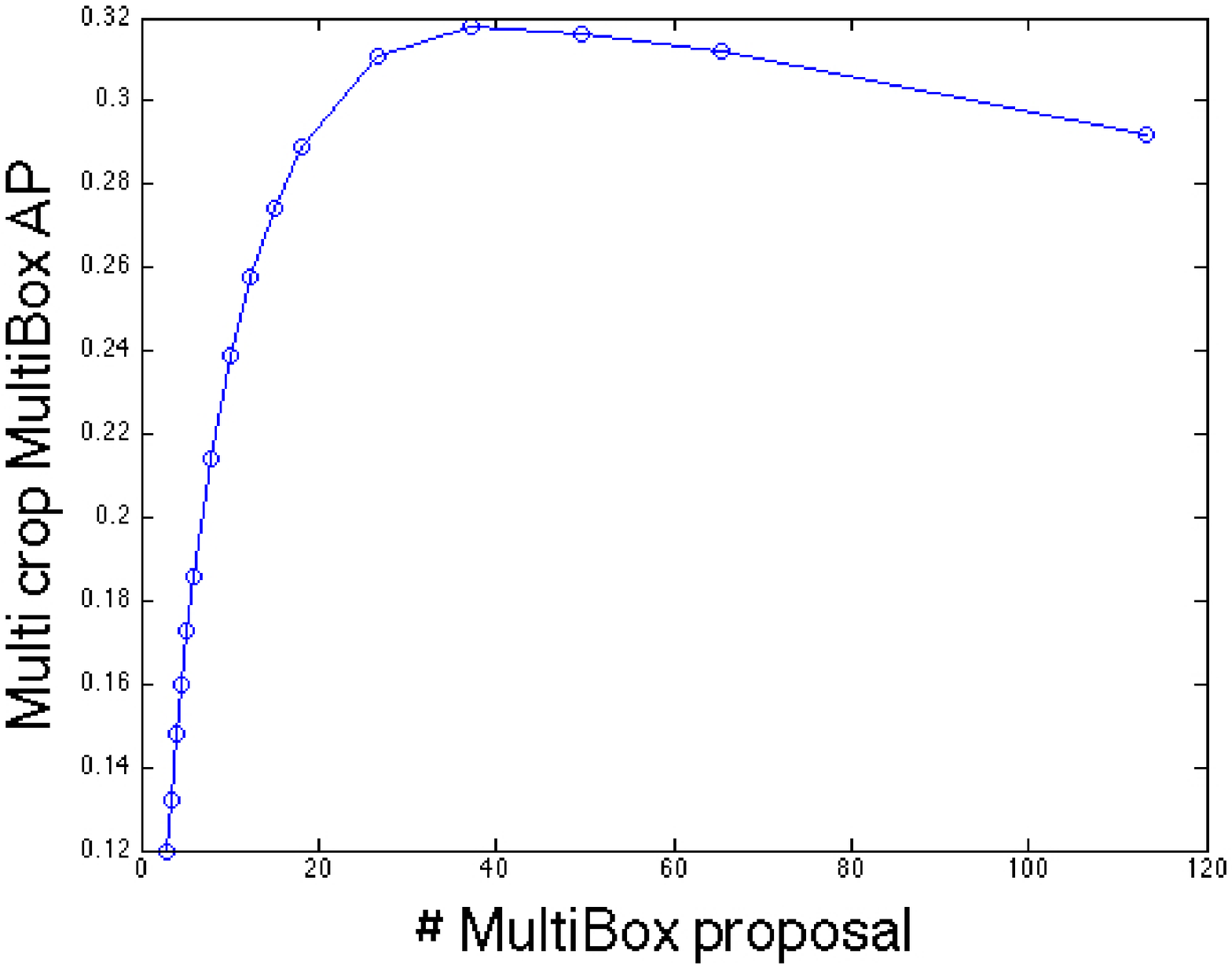}
   }
\end{center}
   \caption{Runtime quality tradeoff.   }
\label{fig:multicrop_ap}
\end{figure*}

Computational efficiency is a major objective for large scale object detection.
In this section, we analyse the trade-off between runtime efficiency and
detection quality. For our approach,  the computational cost is determined
by the number of crops (\ie number of MultiBox evaluations) and the number
of proposals generated by MultiBox (\ie number of post-classification evaluations).
We adopt Average Precision (AP) as the overall quality metric. For automatic
business discovery, it is more important to have high precision results.

Compared to the objects in Imagenet detection task, business store fronts
are relatively small in the entire panorama.  The single crop setting \cite{mbox_2015}
does not apply to our problem, at least not with the 224x224 receptive field
size used for training this network. We have to apply the MultiBox
model at different locations and  different scales, \ie do a multi-crop evaluation.
It is worth noting that the multi-crop evaluation is different from the classic
sliding window approach since the crop does not correspond to the actual
store front.  Figure \ref{fig:multicrop_ap} (a) shows  AP  increases (from
0.304 to 0.358) while the number of crops increases  (from 69 to 904). However,
AP improvement is mostly due to the increase of recall at low precision area.
Figure \ref{fig:multicrop_ap} (b) shows three Precision-Recall curves at
different number of crops. There is a minor performance loss at the high
precision area with fewer crops. 

After MultiBox evaluation, we use a fixed threshold to select proposals for
post-classification. A lower threshold will generate more proposals. Figure
\ref{fig:multicrop_ap} (c) shows performance of AP varies
 as the number of proposals on average per image increases. We select a threshold
that generates  about 37 proposals on
average per panorama, which  gives the best performance. We did
notice that, the performance starts to degrade if we generate too many proposals.

\subsection{Comparison with Selective Search}
 Here we compare with Selective Search in term of both accuracy and runtime
efficiency. We first tuned the Selective Search parameters to get a  max
recall of  62\% with 4666 proposals per image. For comparison, MultiBox achieve
91\% recall with only about 863 proposals per image.  A bigger number of
selective search proposals per image starts to hurt AP.

\begin{figure}[htb]
\begin{center}
   \includegraphics[width=1\linewidth]{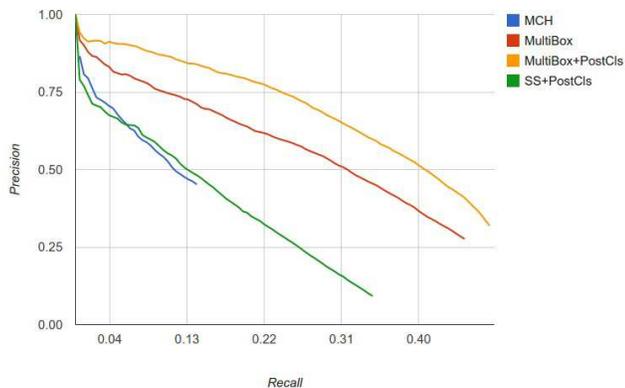}
\end{center}
   \caption{Comparison between MulitBox, MultiBox + Post-Classification,
Selective Search (SS) + Post-Classification and Multi-Context Heatmap (MCH).  }
\label{fig:mbox-ss-comp}
\end{figure}

MultiBox's post-classification model is only trained with MultiBox output
with a low threshold, which  allows MultiBox to propose more boxes to ensure
we have enough negative samples while training post-classiciation. A separate
post-classification model is trained for Selective Search boxes. Both models
are initialized from ILSVRC classification challenge task.  Figure~\ref{fig:mbox-ss-comp}
shows comparison between several approaches. The MultiBox result alone outperforms
Selective Search + Post-Classification with a significant margin. 
  Moreover, the computational cost of our approach is much
lower,  roughly $1/37$  ($=\frac{37+87}{4666}$), than Selective Search+postclassification.
Given a rate of 50 images/second using one network evaluation on a current
Xeon workstation\footnote{Intel(R) Xeon(R) CPU E5-1650 0 @ 3.20GHz, memory
32G.}, this means 2.5 seconds per panorama for our approach as opposed to
1.5 minutes per panorama for Selective Search+post-classification. 

%
\subsection{Comparison with Trained Heat Map Approach}
\begin{figure}[htb]
\begin{center}
   \includegraphics[width=1\linewidth]{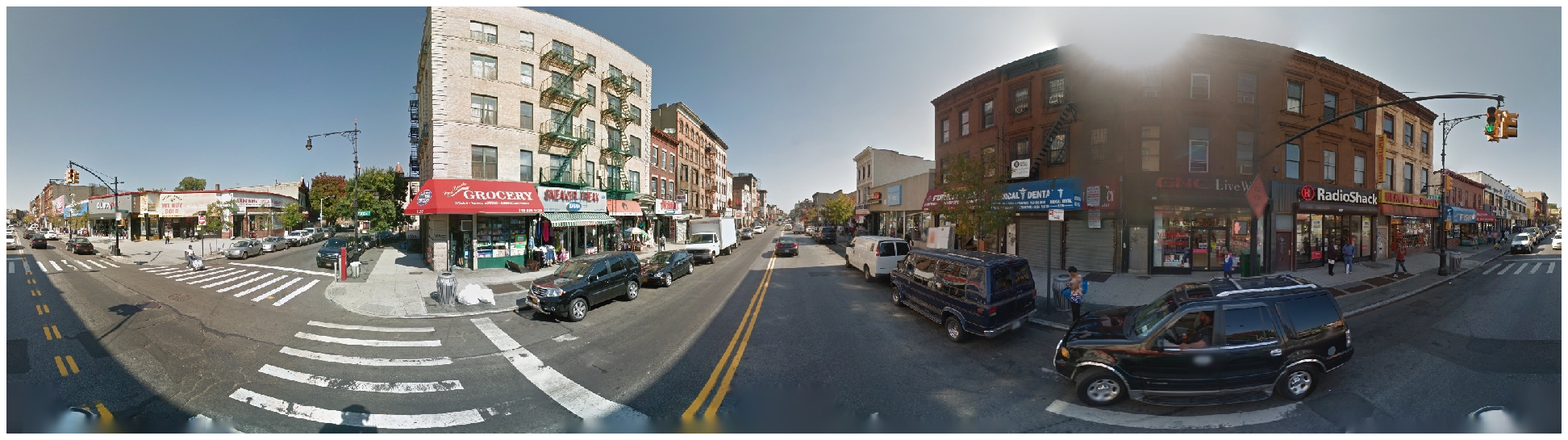}
   (a) Original panorama
   \includegraphics[width=1\linewidth]{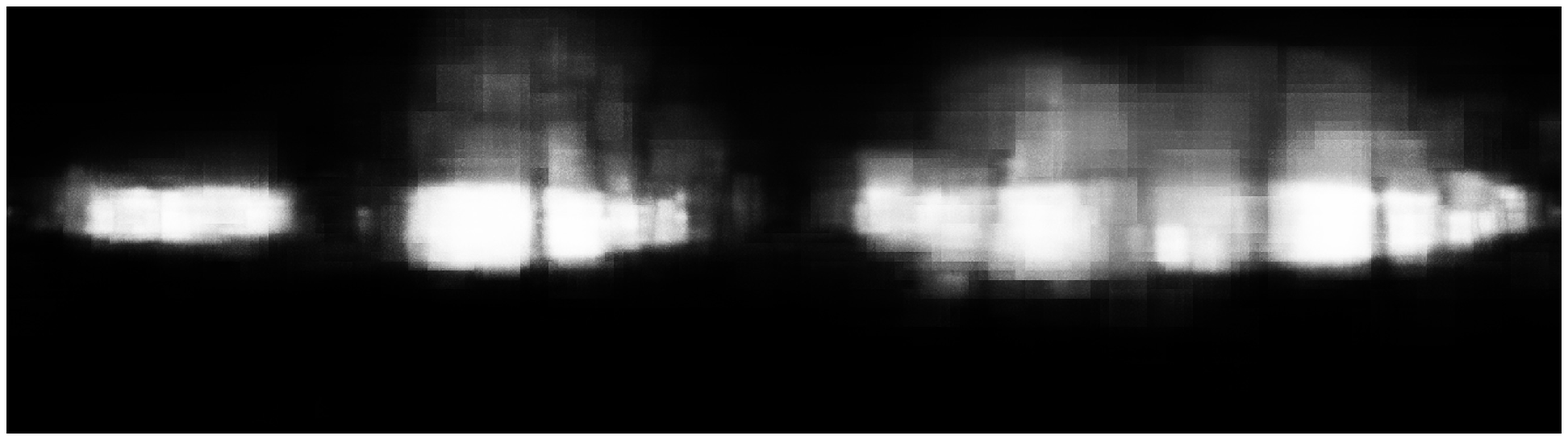}
   (b) Heat map generated from the MCH model   \includegraphics[width=1\linewidth]{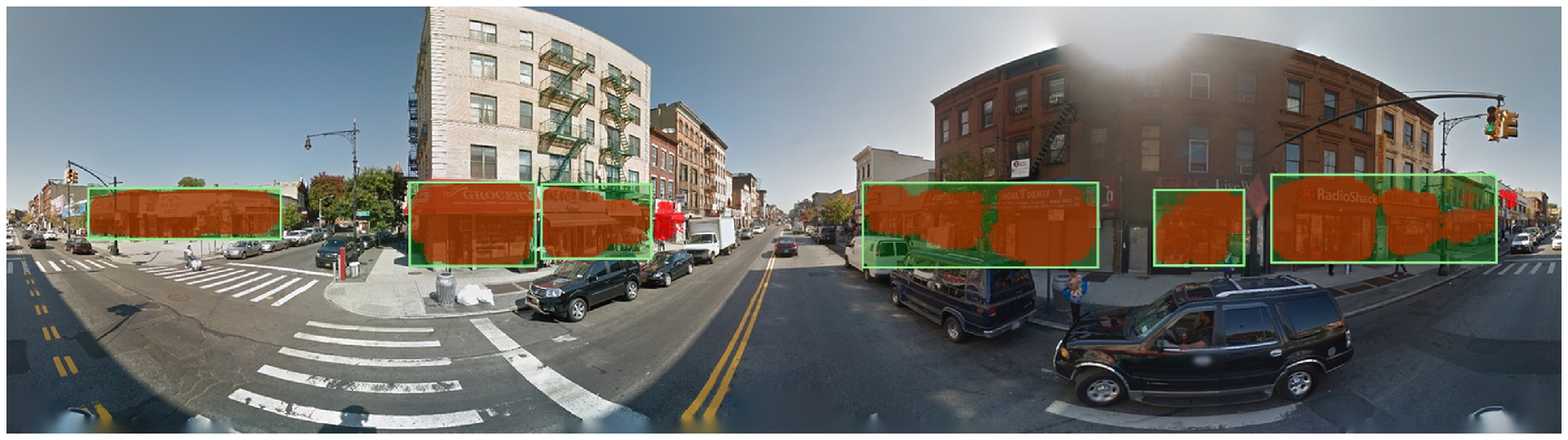}
   (c) Output bounding boxes after postprocessing 
   \includegraphics[width=1\linewidth]{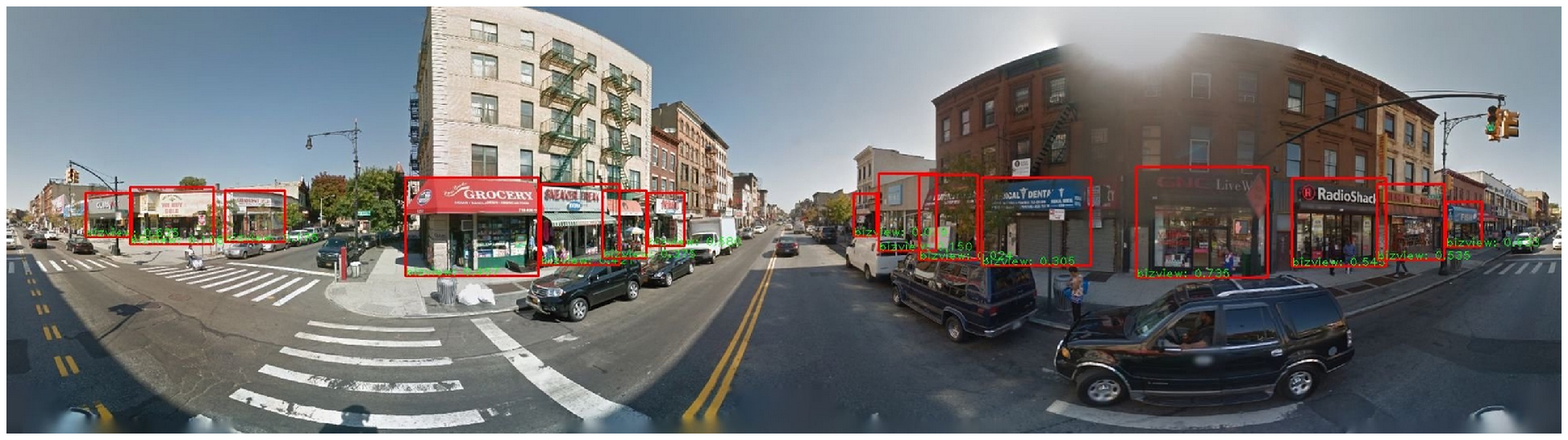}
   (d)
Output from MultiBox\end{center}
   \caption{Compare MultiBox with Multi-Context Heatmap }
\label{fig:hydra-comp}
\end{figure}

In this section, we compare with another Deep Neural Network base object
detection approach in \cite{hydra}, Multi-Context HeatMap (MCH).
Similar to \cite{deep_object}, this approach adopts an architecture that
outputs a heatmap instead of a single classification value.
The main difference of \cite{hydra} compared to \cite{deep_object} is that
it uses a multi-tower convolutional that is fed different resolutions of
the image to get more context information as well as the loss being a simple
logistic regression loss instead of the L2 error minimization proposed in
\cite{deep_object}, which is more discriminative at the pixel level.
A model-free post-processing is used to convert the heat map to detection
results.
This approach has been successfully applied in several detections tasks,
such as text detection, street sign detection where the overall accuracy
of the model was reaching human operator labels or was close to it~\cite{hydra}.
MultiBox significantly outperforms MCH in the business detection case. Figure
\ref{fig:hydra-comp} illustrates the comparison with MCH on one example.
Although the heatmap (Figure \ref{fig:hydra-comp} (b)) generated by MCH is
quite meaningful, converting the heatmap to actual detection windows is a
non-trivial task.
Figure \ref{fig:hydra-comp} (c) shows the detection windows generated by
the post-processing. One can always try different post-processing algorithms.
However, such not learning based post-processing algorithms may either over-segment
or under-segment business store fronts from the heatmap.
Compared to MultiBox, the MCH model is much more sensitive to label noise
present in the training set.
Moreover,  since the cost function is at the pixel level, the MCH model will
have the same penalty on the errors on an border of an object as on the error
far from the border. Thus, it has difficulty to predict the precise boundary.
One reason why MCH works so well on other applications, such as traffic signs
and text, is probably because the boundary definition of these objects is
very well defined and consistent and the objects do not adjoin each other
as business store fronts.

\subsection{Comparison with Human Performance}
Besides the obvious scalability issues, there are a lot of cases in which
human annotators disagree with each other due to ambiguity in business boundaries.
We have sent human annotated store fronts and auto detected store fronts
to human annotators to let them decided whether the box is a store front.
Each question was sent to three different annotators who have been trained
for annotating business store fronts. They did not know if a box has been
generated by the detector or human annotators. A box was confirmed as a true
positive if two or more positive answers were received. We used two different
sets of human annotations: one from the original annotation effort, where
we do not enforce the completeness of the annotation, the other one from
the new annotation effort, where we enforced the completeness of annotation.
We called the first one ``Human Low-Recall'' set and the second one ``Human
High-Recall'' set. The comparison is shown in Table~\ref{tab:human_comp}.
It turns out for both annotation efforts, humans only achieve a precision
below 90\%. In other words, on more than 10\%\ of the annotations human could
not agree with each other. This indicates the ambiguity of labeling the business
store fronts.
Given that humans may  miss annotations as well, it is hard to get true recall.
So we use ``Box Per Image'' as an alternative indicator of coverage. At the
same precision (89.50\%), the  detector already achieves
slightly higher Box Per Image than human annotators in low recall mode. Moreover,
the  detector gives us the flexibility to select an operating point at higher
precision.
\begin{table}[htb]
\begin{center}
\begin{tabular}{|p{3.1cm}|c|c|}
\hline
 & Precision & Box Per Image \\
\hline\hline
Human Low-Recall& 89.50\%\ & \textbf{1.467}\\
\hline
Human High-Recall & 88.72\% & 5.531 \\
\hline
\multirow{2}{*}{\textbf{Auto Detector}}  &  89.50\% & \textbf{1.471}\\
 &  92.00\% & 1.063  \\
\hline
\end{tabular}
\end{center}
\caption{Comparison with human annotation}
\label{tab:human_comp}
\end{table}

\begin{figure*}[ht]
\centering
\subfloat[]{
\includegraphics[height=0.97in]{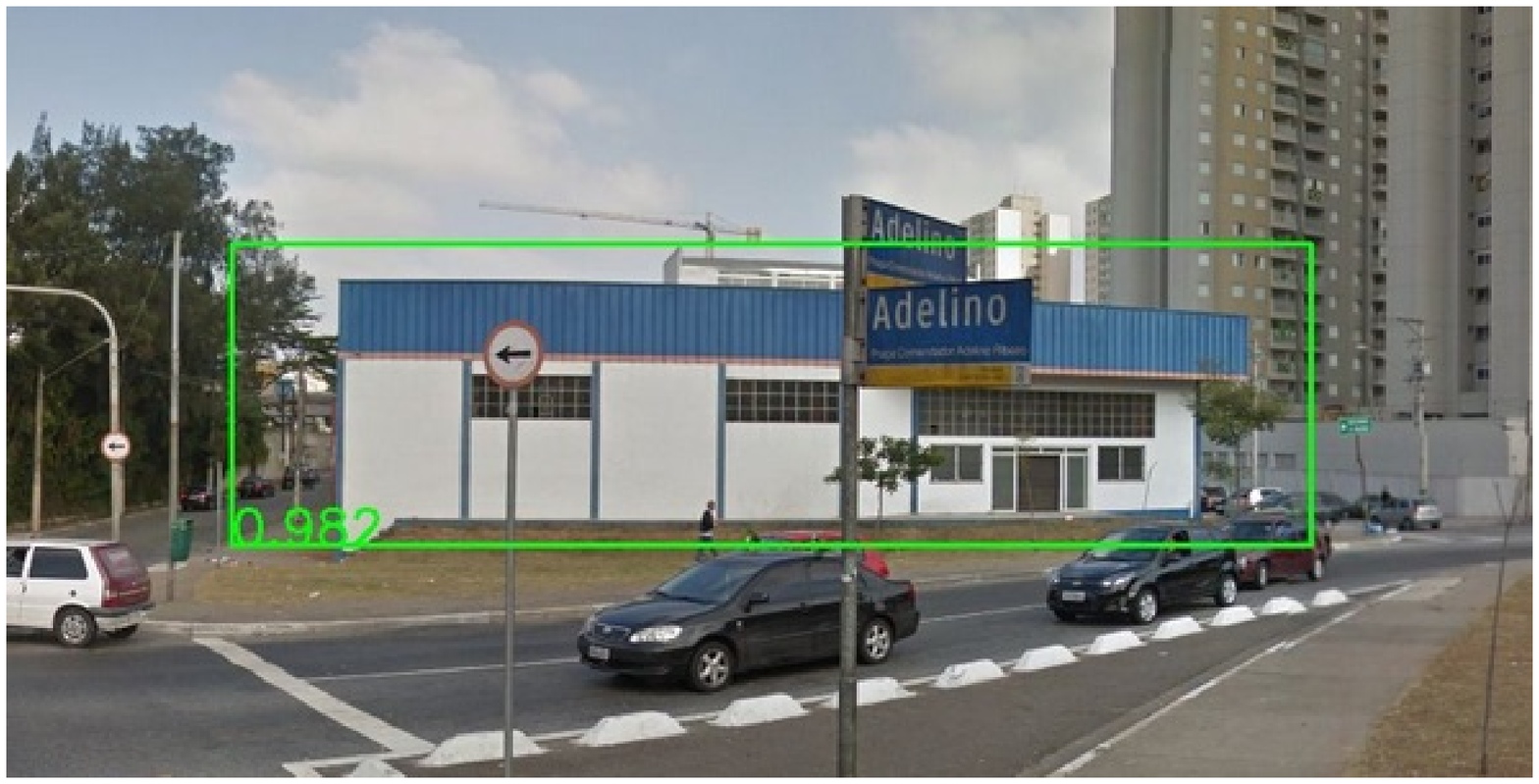} 
}
\subfloat[]{
\includegraphics[height=0.97in]{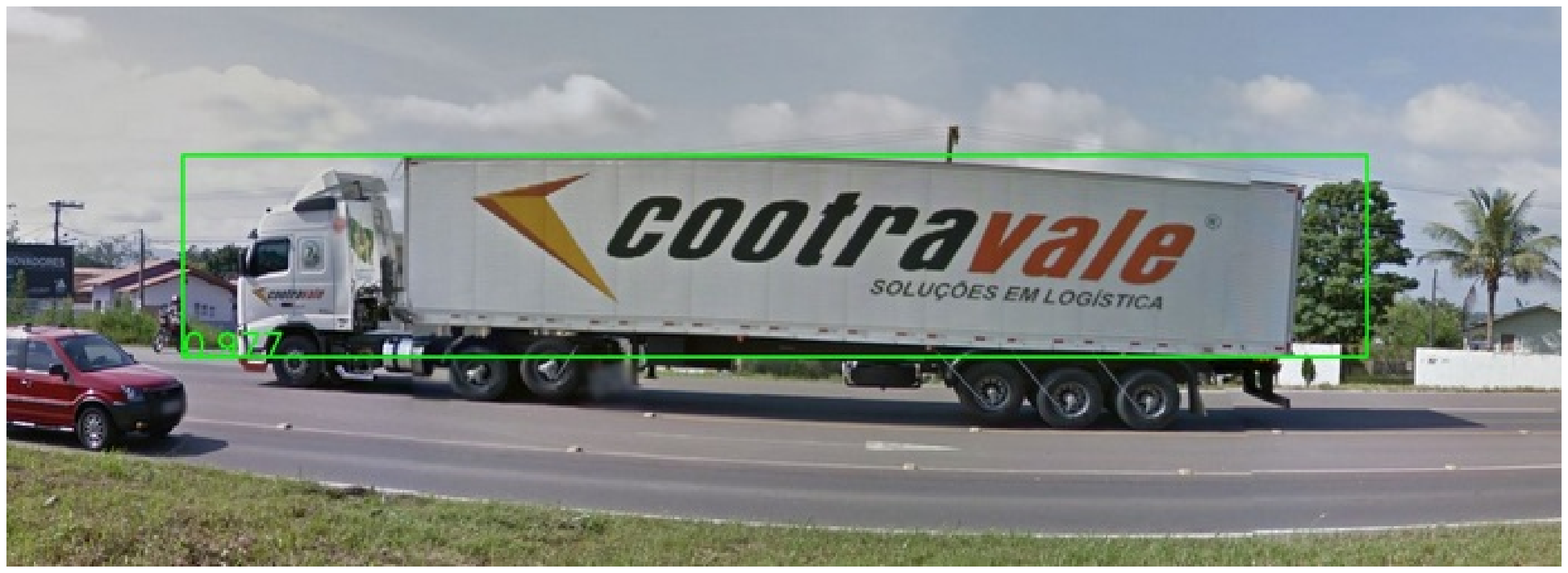} 
}
\subfloat[]{
\includegraphics[height=0.97in]{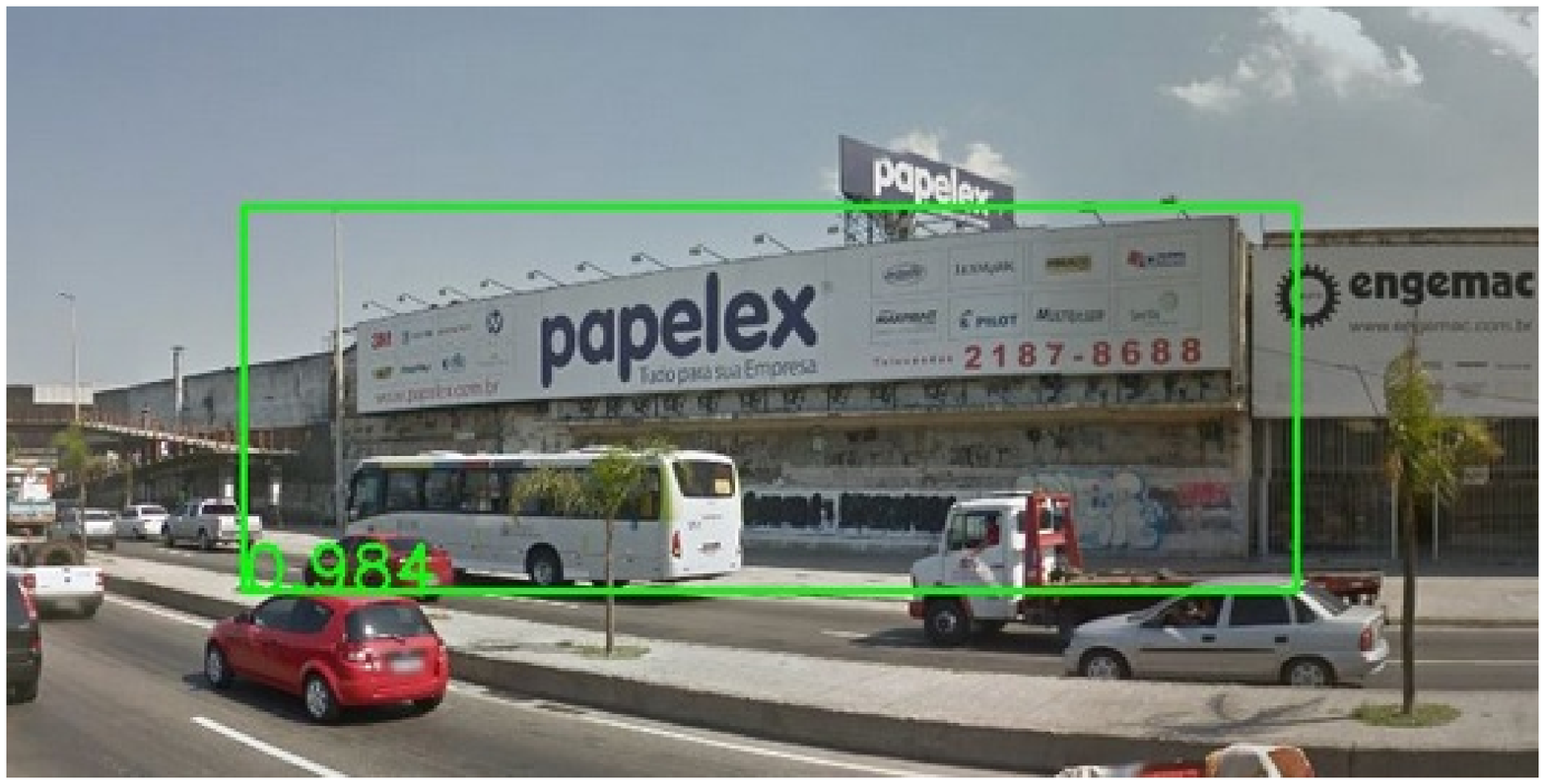} 
}   
\caption{Some typical false positives of detection results}
\label{fig:false_positives}
\end{figure*}

Although the precision of the MultiBox detector is measured to be higher
than that of human annotators at some operating points, we notice that it
tends to generate more egregious false positives than humans did.
Figure \ref{fig:false_positives} shows some of the false positives generated
by the detector. Humans are unlikely
to make mistakes such as in Figure~\ref{fig:false_positives} (a) and (b).
However, business advertisement sign shown in  Figure~\ref{fig:false_positives}
(c) looks so similar to a business store front that can be confusing to human
annotators.
 
\begin{figure*}[tb]
\centering
\includegraphics[width=3.4in]{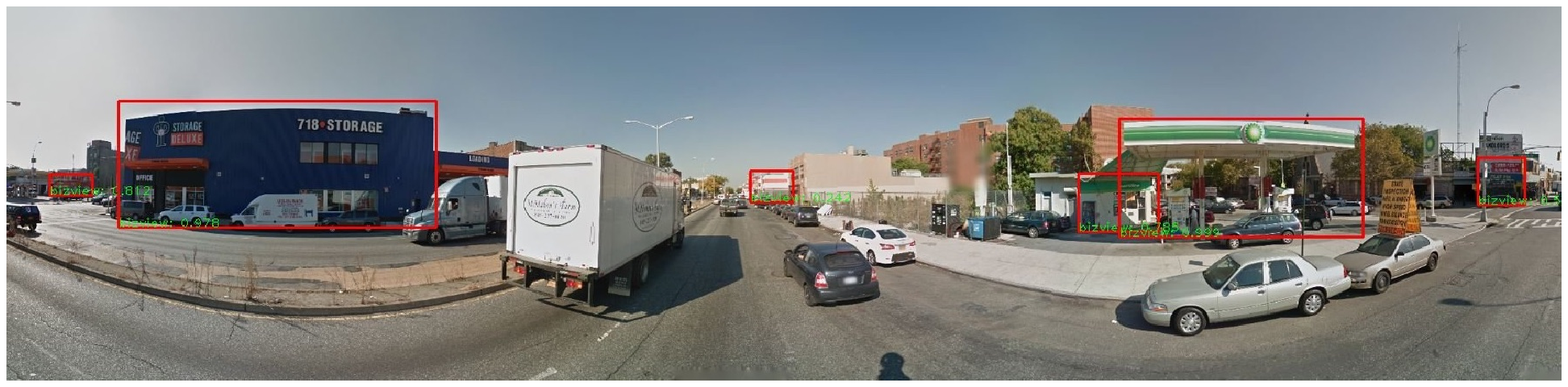}
\includegraphics[width=3.4in]{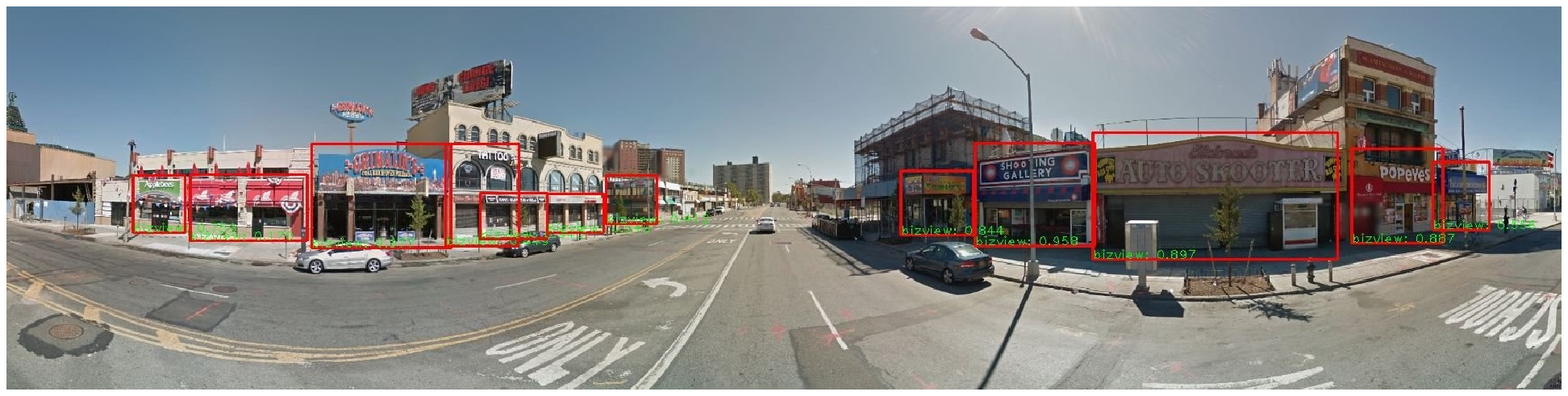}
\includegraphics[width=3.4in]{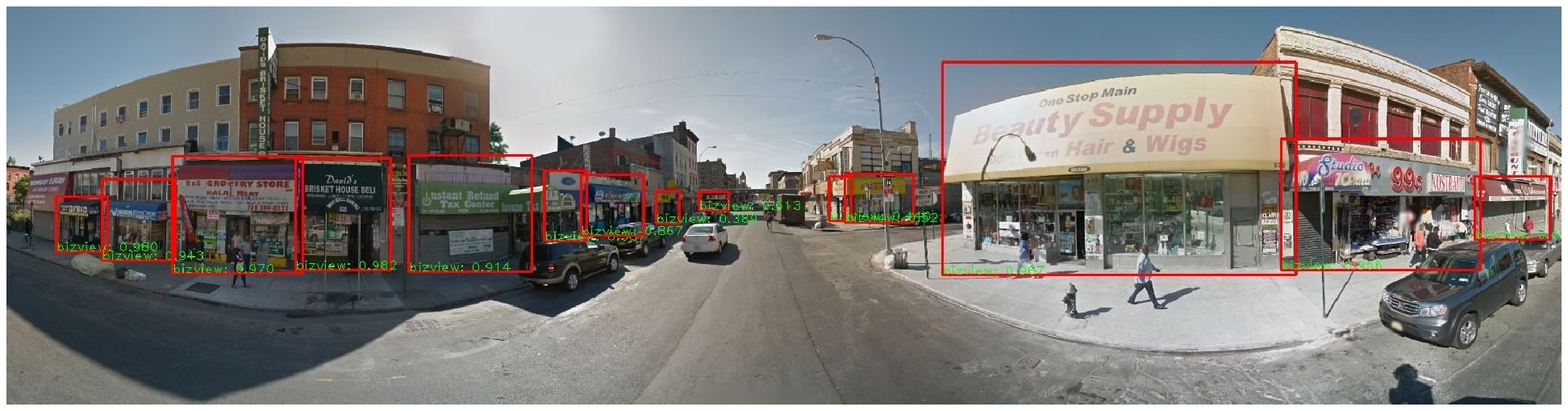}
\includegraphics[width=3.4in]{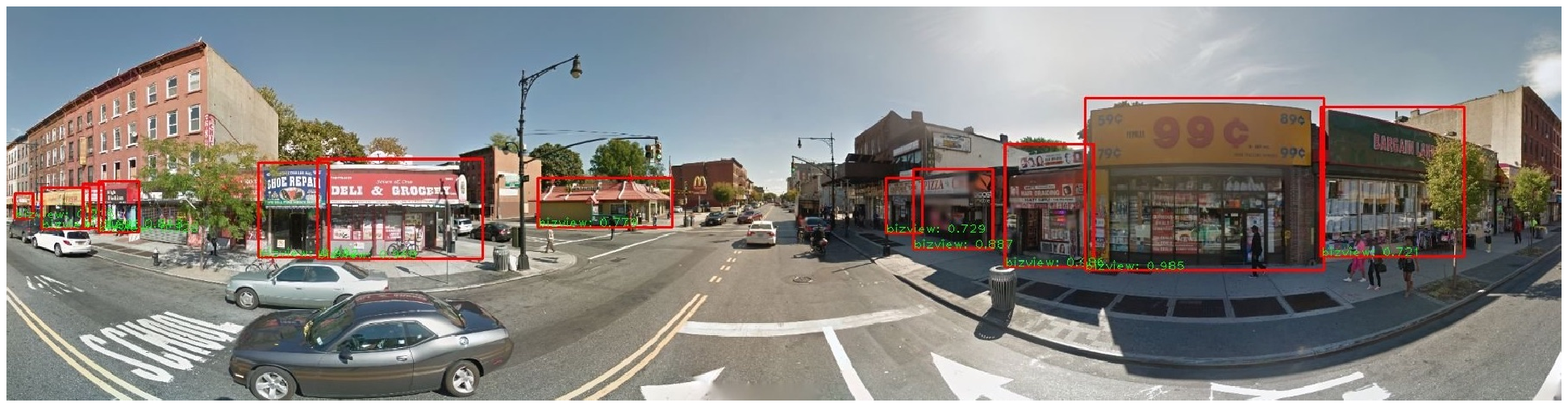}
\caption{Qualitative detection results in panorama space.}
\label{fig:visual-results}
\end{figure*}

\subsection{Business Discovery End-to-End Evaluation}
We used auto detector to generate tens of millions
business store front detections. It is hard to understand actual
business coverage by comparing with the ground truth in the image space.
Given business store front detections from multiple images,
we first merge detections at the same location with a geo-clustering process.
This helps us to remove some of the remaining false positives,
\eg business names on vehicles. GIS information can also help us to further
remove false positives, such as those on high ways, or in residential areas.
Although the complete end-to-end system is out of the scope of this paper,
we would like to provide a better understanding of the
coverage of our automated business discovery process in terms of
precision and recall in the real world.


 For this reason, we conducted a small scale exhaustive end-to-end evaluation
 in Brazil. We selected a metro area of about one square kilometre and
 let annotators have a Street View virtual walk and count all visible businesses.
 In total, 931 unique business were found by this manual process.
 Simultaneously, we let annotators verify the automatically
 detected businesses within the same region before geo-clustering.
 Each automatically detected business in the area was
 sent to three operators and was considered a true positive if
 two or more confirmations were received. The automatic detector
 achieved 94.6\% precision: it got 56 false positives out of the 1045
 detections. Then, we applied geo-clustering to remove the duplicate
 geo-locations from the list of the detections resulting in 495 unique
 businesses. This means a {\bf 53.2\% } recall at {\bf 94.6\%} precision:
 495 out of 931 businesses visible on Street View imagery were correctly
 detected by our automatic system.
 
\section{Summary}
In this paper, we propose to use MultiBox  to
detect business store fronts at scale.
Our approach outperforms two other successful detection techniques
by a large margin.
The computational efficiency of our approach makes the large scale
business discovery worldwide possible. We also compare the detector
performance with human performance and show that human operators
tend to agree more with the detector's output than with human annotations.
Finally, we conducted an end-to-end evaluation to demonstrate the
coverage in physical space.
Given the high computational efficiency of the current detector,
in order to further improve the detector's performance, we will
investigate on using more context features for postclassification and
larger networks.

{\small
\bibliographystyle{ieee}
\bibliography{egbib}
}

\end{document}